\documentclass[lettersize,journal]{IEEEtran}
\usepackage{amsmath,amsfonts}
\usepackage{algorithmic}
\usepackage{algorithm}
\usepackage{array}
\usepackage[font=normalsize]{subfig}
\usepackage{textcomp}
\usepackage{stfloats}
\usepackage{url}
\usepackage{verbatim}
\usepackage{graphicx}
\usepackage{cite}
\usepackage{xpatch}
\usepackage{todonotes}
\usepackage{bm}
\usepackage{multirow}

\makeatletter
\xpatchcmd\algorithmic
  {\newcommand{\IF}}
  {%
    \newcommand\SCOPE{\begin{ALC@g}}%
    \newcommand\ENDSCOPE{\end{ALC@g}}%
    \newcommand{\IF}%
  }
  {}{\fail}
\makeatother

\DeclareMathOperator{\Sim}{sim}
\DeclareMathOperator{\MASK}{MASK}

\graphicspath{{media/}}     % organize your images and other figures under media/ folder

\begin{document}

\title{Spatial HuBERT: Self-supervised Spatial Speech Representation Learning for a Single Talker from Multi-channel Audio}

\author{Antoni Dimitriadis, Siqi Pan, Vidhyasaharan Sethu, Beena Ahmed
        % <-this % stops a space
\thanks{Antoni Dimitriadis, Vidhyasaharan Sethu and Beena Ahmed are with The School of Electrical Engineering \& Telecommunications, UNSW Australia, Sydney, Australia (email: antoni.dimitriadis@unsw.edu.au, v.sethu@unsw.edu.au, beena.ahmed@unsw.edu.au)}% <-this % stops a space
\thanks{Siqi Pan is with Dolby Laboratories, Sydney, Australia (email: siqi.pan@dolby.com)}}

% \markboth{}


\maketitle

\begin{abstract}
Self-supervised learning has been used to leverage unlabelled data, improving accuracy and generalisation of speech systems through the training of representation models. While many recent works have sought to produce effective representations across a variety of acoustic domains, languages, modalities and even simultaneous speakers, these studies have all been limited to single-channel audio recordings. This paper presents Spatial HuBERT, a self-supervised speech representation model that learns both acoustic and spatial information pertaining to a single speaker in a potentially noisy environment by using multi-channel audio inputs. Spatial HuBERT learns representations that outperform state-of-the-art single-channel speech representations on a variety of spatial downstream tasks, particularly in reverberant and noisy environments. We also demonstrate the utility of the representations learned by Spatial HuBERT on a speech localisation downstream task. Along with this paper, we publicly release a new dataset of 100 000 simulated first-order ambisonics room impulse responses.
\end{abstract}


\section{Introduction}
Speech, as one of the most fundamental forms of human communication, carries a wealth of information, ranging from linguistic content to emotional cues and speaker characteristics. Inspired by the human brain, the goal of a speech representation learning (SRL) model is to extract this information in a way where it can be readily accessed by the simplest of downstream models, even in the presence of complex, structured noise sources that overlap with the target speech \cite{Bengio2013_RLreview}. Unlike the human auditory system however, current speech representation models view speech as a single-channel audio signal, and are unable to utilise the rich spatial information that is present in multi-channel audio. This spatial information enables humans to both track the location of speech sources in space, and also to better isolate them from many forms of interfering noise. As the majority of modern commercial devices such as mobile phones and smart speakers contain multiple microphones, the ability to exploit this spatial information through the representation learning process has the potential to lead to significant improvements in performance when building speech processing systems for these devices.

Despite lacking multi-channel capabilities, representation learning techniques have shown significant promise when applied to speech signals, and offer many benefits over training end-to-end systems. Early approaches used supervised pre-training \cite{Zeiler2014_cnns}, sometimes referred to as transfer learning \cite{Wang2015_transferspeech}. Supervised pre-training optimises a model to solve a specific downstream task on a large labelled dataset, and then re-uses the learned weights either for new tasks, or on new datasets \cite{Mahajan2018_cnnslimits}. In recent years however, significant progress has been made in the field of speech representation learning through the use of self-supervised learning (SSL), with the development of models such as wav2vec2.0 \cite{Baevski2020_wav2vec2}, HuBERT \cite{Hsu2021_HuBERT} and WavLM \cite{Chen2021_WavLM}. Unlike supervised pre-training methods, self-supervised pre-training does not require the use of external labels. Instead, a proxy task is designed that extracts training labels from the input data itself. These proxy tasks typically involve predicting unseen information extracted from future frames in the sequence, or frames that are masked to the model input, and can use regression, classification, or contrastive losses \cite{Mohamed2022_SSSRLReview}.

The major advantage of self-supervised pre-training is the ability to leverage large amounts of unlabelled data, allowing the models to train on multiple domains and covering a wide variety of conditions. This results in representations that generalise well to out of domain data, with far less performance degradation when evaluating on domains unseen during training \cite{Hsu2021_RobustWav2vec2, Zuluaga-Gomez2023_OOD-ASR}. Supervised pre-training objectives encourage models to discard information not needed for the pre-training task, while due to the lack of labels, representations learned from self-supervised objectives are more universal than those trained in supervised settings \cite{Hsu19_transferAVG, Jia2018_transferSV, Chen2021_SpeechNet}, and can achieve reasonable performance on a wide range of downstream tasks \cite{Yang2021_SUPERB, Tsai2022_SUPERBSG}. Building general purpose pre-trained models for speech enables significant improvements in tasks with limited access to supervised training data.

Self-supervised speech representation models have also enabled several completely novel applications such as unsupervised speech recognition \cite{Baevski2021_USR} and synthesis \cite{Lie2022_USSSynth}. Previous studies have also extended these representations to multi-lingual data \cite{Kawakami2020_multilingual, Conneau2021_UScross-lingual, Babu2022_xls-r}, multi-modal data \cite{Shi2022-avhubert,Hsu2022_uhubert}, and recently mixtures of multiple speakers \cite{Fazel-Zarandi2023_cocktailhubert}, all showcasing the benefits of training speech representations in a wide variety of downstream scenarios.

%\IEEEpubidadjcol

Despite the significant progress made in these works, these models are all restricted to to single-channel recordings in which the target speaker is typically in close proximity to the microphone. In order to retain the benefits of these representation models and still exploit the multi-channel capabilities of many recording devices, modern speech processing systems must use either classical signal processing techniques or separately trained non-linear models to first perform multi-channel speech enhancement in order to extract a de-noised single channel speech signal to pass to a representation model \cite{Markovich-Golan2018, Erdogan2016_ImprovedMVDR}. However, these systems are designed to remove the spatial information from the input signal making it completely inaccessible to downstream models. Instead, we seek to build a new self-supervised speech representation model directly from multi-channel inputs, allowing for both cleaner representations in the presence of spatial noise sources, and also enabling downstream models to directly access spatial information for tasks such as speaker localisation.

In this paper, we introduce Spatial HuBERT (Sp-HuBERT), a self-supervised training framework that pre-trains on simulated multi-channel recordings of reverberant speech. Sp-HuBERT follows the masked speech prediction and denoising framework used in WavLM \cite{Chen2021_WavLM}, with the addition of a masked spatial prediction loss. Training effective speech representations requires a large training corpus, far more than any publicly available multi-channel speech datasets. To combat this issue, Sp-HuBERT utilises simulated room impulse responses in the first-order ambisonics domain to convert large single-channel datasets into a suitable format for self-supervised pre-training.

We compare our model to the state-of-the-art single channel speech representation of a similar size, WavLM Base+, on a selection of tasks from the SUPERB Benchmark \cite{Yang2021_SUPERB} converted to a spatial audio format. In noisy and reverberant conditions, Sp-HuBERT achieves a relative reduction of over 40\% in word error rate on Librispeech over WavLM Base+, despite using nearly 100 times less data for pre-training.

We implement our upstream model and training process using the Fairseq toolkit \cite{Ott2019_fairseq}, and implement our downstream evaluation tasks using the s3prl toolkit \cite{Yang2021_SUPERB, Tsai2022_SUPERBSG}. Along with our code, we release a new dataset of 100 000 simulated FOA impulse responses \footnote{FOA IR Dataset hosted on Huggingface: \url{https://huggingface.co/datasets/adimitri/sp-hubert_impulse_responses}}.

The rest of this paper is organised as follows. Section~\ref{sec:related_work} highlights some key related publications on which our work is based. Section~\ref{sec:background} gives a brief technical overview of the Ambisonics spatial format, and the Masked Prediction Loss utilised in our work. Section~\ref{sec:sp-hubert} details the Sp-HuBERT architecture, losses and data augmentation techniques. Section~\ref{sec:experimental_setup} provides experimental details including all hyper-parameter values used for training both our upstream model, and all of the downstream models. Section~\ref{sec:results} presents our results, including experiments detailing how performance varies in noisy and reverberant conditions.

\section{Related Work}\label{sec:related_work}
This work builds upon two existing single-channel speech representation learning models, Hidden-Unit BERT (HuBERT) \cite{Hsu2021_HuBERT} and WavLM \cite{Chen2021_WavLM}. The HuBERT architecture is made up of two main blocks: the first block consists of several CNN layers that down-sample the input into frames with a stride of 20ms, and the second block is a stack of transformer encoders that are able to use utterance-wide context to learn deep representations of the speech. HuBERT introduces a novel self-supervised learning objective, masked prediction loss, heavily inspired by the Masked Language Modelling loss used by the BERT language model. HuBERT uses unlabelled clean speech recordings to pre-train the speech representation model for use on an automatic speech recognition (ASR) downstream task. We describe this loss in more detail in section~\ref{sec:MCP}. While the BERT language model uses the input token itself as the label, HuBERT obtains discrete pseudo-labels for each frame via a K-means clustering of audio features. The HuBERT model initially trains on labels generated by clustering mel-frequency cepstral coefficients (MFCCs), and later generates new labels using features from the 6th layer of its transformer encoder.

WavLM expands on the HuBERT framework with some small modifications to the transformer architecture by replacing the absolute position bias with a gated relative position bias \cite{Chi2022_XLME}, and additionally introducing a denoising component to the training process. Rather than training on clean speech, WavLM mixes utterances with randomly sampled within-batch secondary speech, or with recorded noise samples taken from the Deep Noise Suppression Challenge dataset \cite{Reddy2021_DNS}. These changes lead to improved overall performance on a variety of downstream speech tasks, with particular improvement on speaker identification.

Additionally, our downstream evaluation methodology is based heavily upon the Speech Universal PERformance Benchmark (SUPERB) \cite{Yang2021_SUPERB}. The SUPERB Challenge consists of a broad set of speech processing tasks, each with a prescribed downstream model architecture, and compares speech representation models by evaluating their performance on each task without fine-tuning. Tasks are selected to cover the diverse range of information present in speech signals, and are categorised as either speaker, content, semantic, para-linguistic, or generative.

\section{Background}\label{sec:background}
\subsection{Higher Order Ambisonics Format}
Higher Order Ambisonics (HOA) is a \textit{system-independent} spatial audio format used for capture and reproduction of sound in a full three-dimensional sphere \cite{Gerzon1973_Periphony}. HOA represents the sound-field as a series of spatially-orthogonal spherical harmonics. Multi-channel microphone signals from any fixed array configuration with enough channels can be converted into HOA components by computing the weighted scalar products between the signals and the corresponding spherical harmonic functions for each channel \cite{Kitic2018_TRAMP}. A continuous sound-field can be reproduced as an infinite linear combination of these so-called HOA components with high accuracy \cite{Ward2001_HOAReproduction}. 

In practice, the representation is truncated to a desired order, and only a fixed number of HOA components are used. First-order Ambisonics (FOA), is the first-order truncation of HOA, consisting of 4 channels, typically referred to as W (omnidirectional), X (front-to-back), Y (left-to-right), and Z (up-and-down).

\subsection{Masked Prediction Loss}\label{sec:MCP}
Similarly to the language model BERT \cite{Devlin2019_BERT}, the Masked Prediction training objective masks a portion of the input sequence and trains the model to predict a label associated with each of the masked frames from the context of surrounding unmasked frames. More formally, let $\bm x$ be a speech waveform, $\bm y = [y_1,\dots,y_T] = f_t(\bm x)$ be the output of the CNN-block, $h_t = g_t(\bm y)$ be the output of the of the $L$-layer transformer encoder block at time $t$, and $z_t$ be the class-label for the frame at time $t$. The model parameterises the distribution over the classes as
\begin{equation}\label{eq:CE_dist}
    p(c\,|\, \bm y, t) = \frac{\exp(\Sim(Ag_t(\bm y),e_c)/\tau)}{\sum_{c'=1}^{C}\exp(\Sim(Ag_t(\bm y),e_{c'})/\tau)},
\end{equation}
where $c \in [1,C]$ is the true class label of frame $t$, $A$ is a trainable projection matrix, $e_c$ is the trainable embedding for class $c$, $\Sim(a,b)$ computes cosine similarity, and $\tau$ is a logit scaling factor that we set to 0.1 as in prior works. The masked prediction loss is given by
\[\mathcal{L} = \sum_{t\in \mathcal{M}} -\log p(z_t \, | \, \MASK(\bm y), t), \]
where $\MASK(\cdot)$ randomly replaces frames with a trainable masked embedding, and $\mathcal{M}$ is the set of all frames that are masked.

\section{Spatial HuBERT}\label{sec:sp-hubert}
We present Spatial HuBERT (Sp-HuBERT), a multi-channel self-supervised speech representation model trained to produce noise-robust speech representations using room impulse responses for a fixed spatial configuration. We extend the single-channel training objectives used by WavLM with spatial audio simulation. By using simulated spatial audio, our training data is not restricted by the limited availability of multi-channel recordings.

\subsection{Simulating Spatial Data}\label{sec:spatialisation}
It is necessary to assume a fixed microphone array configuration at the input to the model. In order to maximise the adaptability, we selected the First-order Ambisonics (FOA) format. FOA is a full-sphere system-independent format, and only requires 4 channels at the input. Recordings from different microphone array configurations can be converted into FOA if necessary, but larger arrays may lose some spatial resolution in the process, and planar arrays will have no resolution in the perpendicular axis.

While there are some publicly available FOA impulse response datasets \cite{Politis2022_TAU}, they are insufficient in size for self-supervised learning. We utilise a statistics-based impulse response (IR) generation algorithm to produce a large dataset of FOA impulse responses. IR properties are controlled by specifying room dimensions (height, width, and length), source location, and RT60 parameters. In lieu of releasing the code used for IR generation, we release the dataset of 100 000 simulated impulse responses, generated using parameters given in table~\ref{tab:ir_params}.

\begin{table}[t]
    \centering
    \begin{tabular}{|c|c|c|} \hline
        \textbf{Parameter} & \textbf{Description} & \textbf{Distribution} \\ \hline
        $L$ & Room Length                      & $L\sim U(3,6)$ \\ \hline
        $W$ & Room Width                       & $W \sim U(2,5)$ \\ \hline
        $H$ & Room Height                      & $H \sim U(3,4)$ \\ \hline
        $x$ & \multirow{3}{*}{Source Location} & $x \sim U(0.5,L)$ \\
        $y$ &                                  & $y \sim U(0.5,W)$ \\
        $z$ &                                  & $z \sim U(0.5,H)$ \\ \hline
        RT60 & Reverberation Time              & RT60 $\sim N(0.45,0.18)$ \\ \hline
    \end{tabular}
    \vspace{2mm}
    \caption{Table of parameters used for IR generation}
    \label{tab:ir_params}
\end{table}

We convert clean single-channel speech recordings into stationary FOA spatial speech by convolving with the generated impulse responses. Specifically, given a clean speech recording $\bm a$ of length $L$ samples, and an impulse response $\bm{u}$ with a direction label $l$ we set
\[ \bm a' = \bm a\ast \bm u, \quad \bm l = 
\bigl( \underbrace{%
    l, l, \cdots, l
  }_{\text{$L$~elements}} \bigr), \]
where $\bm a'$ is the simulated multichannel speech, and $\bm l$ is a sequence of DOA labels for each frame. Our impulse response generation method could not be easily extended to the case of moving sound sources. As a result, we simulate moving sound sources in a free field environment (no reverberation) by computing the FOA gains at each position along the trajectory. 

We limit our simulations to linear trajectories, and restrict the velocity of the potential source. Specifically, with maximum initial distances of the source to the microphone array of $m_x, m_y, m_z$ along the $x,y,z$ axes respectively, a minimum distance from the microphone array $m_\text{dist}$ we first randomly sample $x \sim \mathcal{U}(-m_x,m_x), y \sim \mathcal{U}(-m_y,m_y), z \sim \mathcal{U}(-m_z,m_z)$ such that $||(x,y,z)|| > m_\text{dist}$, and set our start point $s = (x,y,z)$. Next, we randomly sampled a trajectory length $|d| \sim \mathcal{U}(0, Lv_\text{max}/f_s)$, where $v_\text{max}$ is the maximum source velocity. The trajectory direction is uniformly sampled on the surface of a unit sphere using the rejection method. That is, we sample $d_x,d_y,d_z \sim \mathcal{U}(-1,1)$ until $||d_x,d_y,d_z|| \le 1$, and set the trajectory direction 
\[\overrightarrow{d} = \frac{(d_x,d_y,d_z)}{||d_x,d_y,d_z||}.\]
We also reject samples where the trajectory extending from $s$ along these direction will pass within $m_\text{dist}$ meters of the microphones. That is, if
\[ \frac{||s\times\overrightarrow{d}||}{||\overrightarrow{d}||} > m_\text{dist} \]
then we reject and re-sample the trajectory direction $\overrightarrow{d}$. We set the trajectory end-point $e = s + |d|\cdot\overrightarrow{d}$, and the full sampled trajectory at each sample $i$ is given by 
\[g_i = e\cdot \frac{i-1}{L-1} + s\cdot \frac{L-i}{L-1}\] 
for each $i$ from $1$ to $L$. The normalised direction label at sample $i$ can be obtained from the trajectory as 
\[ l_i = \frac{g_i}{||g_i||}.\]
Finally, our spatial audio source at sample $i$ is assigned the values
\[ a'_i = \frac{a_i d_{\min}}{||g_i||} (1, l_{i_x}, l_{i_y}, l_{i_z} ) \]
where $\bm a = (a_1,\dots,a_L)$ is a clean single-channel recording, $l_{i_x}, l_{i_y}, l_{i_z}$ are the normalised $x,y,z$ coordinates of the source at sample $i$, $d_{\min} \gets \min_i{(||g_i||)}$ is the closest point on the trajectory to the microphone array, and $||g_i||$ is the distance of the source to the array at sample $i$. The W channel simply receives the original recording, while the X, Y, and Z channels at each sample are multiplied by the normalised co-ordinates of the source. The scaling factor of $d_{\min}/||g_i||$ accounts for the change in intensity due to the change in distance between the source and microphones.

Our training data is made from a mixture of reverberant, stationary simulated sound sources using the generated impulse responses, and free field, moving sound sources simulated using the method described above. The proportion of the mixture is controlled with a fixed ratio $p_r$. During training, with probability $p_r$ we select the stationary source approach, and with probability $1-p_r$ we select the moving source approach.

\subsection{Model Architecture}
\begin{figure}[t]
    \centering
    \includegraphics{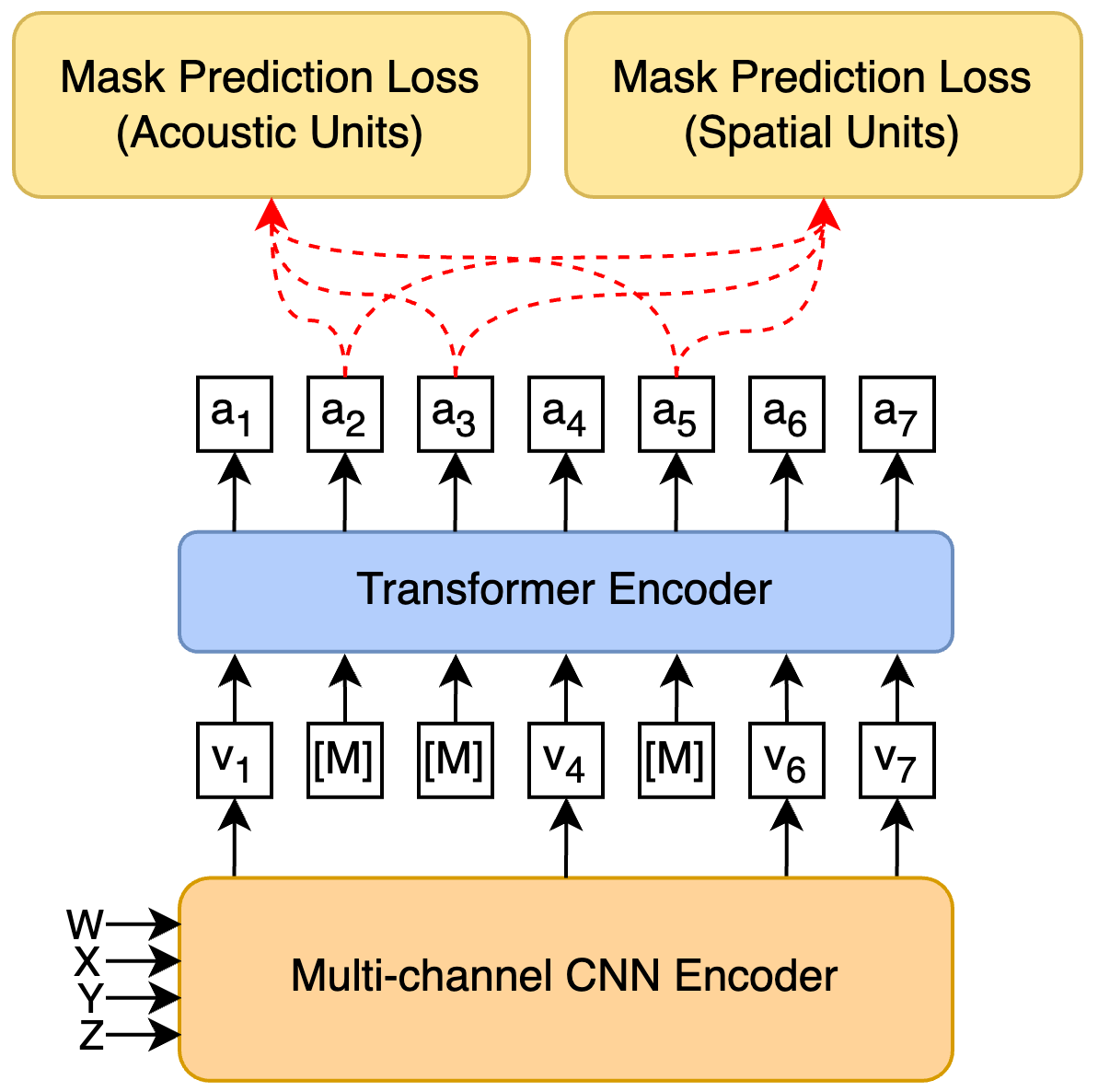}
    \caption{Sp-HuBERT model architecture}
    \label{fig:architecture}
\end{figure}

Figure~\ref{fig:architecture} shows the overall model structure for Sp-HuBERT. Similarly to single-channel speech representations, the Sp-HuBERT model architecture consists of a convolutional feature encoder followed by a transformer encoder. The convolutional encoder takes a 4-channel input, and is built of 7 layers of temporal convolutions followed by a layer normalisation. Each layer has 1024 channels and uses a GELU activation \cite{Hendrycks2023_GELU}, with strides of (5,2,2,2,2,2,2) and (10,3,3,3,3,2,2) respectively, resulting in frames of approximately 25ms wide with a 20ms stride. Sp-HuBERT uses double the channel count of WavLM in each convolutional layer, to allow the encoder to represent cross-terms between channels in the input.

The transformer encoder uses the same structure as WavLM Base. It is comprised of 12 transformer layers, each with 12 attention heads and 768-dimensional hidden states, and utilises a gated relative position bias on the first layer.

\subsection{Training Objective}
As shown in figure~\ref{fig:architecture}, Sp-HuBERT utilises a two-part masked prediction loss, as described in section~\ref{sec:MCP}. The first part aims to learn acoustic units by using pseudo-labels generated by K-means clustering the 6th layer of a 1st iteration trained HuBERT model, similarly to both the HuBERT Base model and the WavLM Base model.

In addition to the acoustic loss, there is also a spatial loss component to encourage learning spatial information. The spatial loss uses quantised direction labels generated from direction-of-arrival (DOA) information available from the spatialisation process described in section~\ref{sec:spatialisation}. DOA labels for each frame are converted into azimuth and elevation angles, and discrete labels are generated by a uniform segmentation in each dimension. Specifically, for frame $t$ with a normalised position $(x,y,z)$, we assign it a discrete label $\zeta_t$ as
\[ \zeta_t = \left\lfloor\frac{n\theta}{\pi}\right\rfloor + n\left\lfloor\frac{m\phi}{2\pi}\right\rfloor \]
where $\theta = \arccos(z)$ is the elevation of the source ranging from 0 to $\pi$, $\phi = \arctan(y,x) + \pi$ is the azimuth of the source ranging from 0 to $2\pi$, $n$ is the number of segments in elevation, and $m$ is the number of segments in azimuth. This results in a total of $nm$ discrete classes for the classification task.

The total loss is a weighted sum of these two components. Specifically, 
\begin{align*}
\mathcal{L}_\text{acoustic} &= \sum_{t\in M} -\log p(z_t \, | \, \MASK(\bm y), t) \\
\mathcal{L}_\text{spatial} &= \sum_{t\in M} -\log p(\zeta_t \, | \, \MASK(\bm y), t) \\
\mathcal{L}_\text{total} &= \mathcal{L}_\text{acoustic} + \lambda\mathcal{L}_\text{spatial} 
\end{align*}
where $z_t$ and $\zeta_t$ are the acoustic and spatial class labels respectively for frame $t$, $p$ is defined as in equation~\ref{eq:CE_dist}, and $\lambda$ is a hyper-parameter that adjusts the weight of the spatial loss.

Sp-HuBERT also makes use of data augmentation akin to WavLM by mixing DNS noise and secondary speech into utterances during training. A similar utterance mixing protocol to WavLM \cite[Alg. 1]{Chen2021_WavLM} is employed. For each batch of spatial speech signals, utterances are mixed with some probability $p_m$. If mixing occurs, the interfering signal will be sampled from a DNS noise dataset with probability $p_n$ and spatialised using the method given in section~\ref{sec:spatialisation}, or otherwise sampled from a secondary speech utterance from within the same batch. If the interference is speech, it is truncated to be at most half the length of the primary signal. The primary speech is mixed with the interference at a random selected SNR.

% \begin{algorithm}[H]
% \caption{Noisy/Overlapped Spatial Speech Simulation}\label{alg:utt_mixing}
% \begin{algorithmic}
% \STATE {\textbf{Given:}} A batch of speech utterances $\bm X = \{\bm x_i\}_{i=1}^B$ with batch size $B$ and length $L$, reverberation probability $p_r$, a set of DOA labelled impulse responses $\bm U = \{\bm u_i\}_{i=1}^K$ of size $K$, mixing probability $p_m$, noise probability $p_n$, and a set of DNS noise recordings $\bm N = \{\bm n_i\}_{i=1}^M$ of size $M$.
% \vspace{1mm}
% \STATE {\textsc{Mix}}$(\bm X, p_r, \bm U, p_m, p_n, \bm N)$:
% \SCOPE
% \STATE{$\bm x'_i$ = \textsc{Spatialise}$(\bm x_i, p_r, \bm U)$ \textbf{ for } $ i \in 1,B $ }
% \FOR{each $\bm x' \in \bm X'$}
% \STATE{Randomly sample $v \sim \mathcal{U}(0,1)$}
% \IF{$v < p_m$}
%     \STATE{Randomly sample $v \sim \mathcal{U}(0,1)$}  
%     \IF{$v < p_n$}
%         \STATE Randomly sample noise $\bm n$ from $\bm N$.
%     \ELSE
%         \STATE Randomly sample a secondary speech utterance $\bm n$ from the batch $X$ with uniform probability.
%     \ENDIF
%     \STATE Sample SNR $\sim \mathcal{U}(-5,20)$.
%     \STATE Sample mixing length $l \in \{ \}$
% \ENDIF
% \ENDFOR
% \ENDSCOPE
% \end{algorithmic}
% \end{algorithm}

\section{Experimental Setup}\label{sec:experimental_setup}
\subsection{Upstream Training}\label{sec:upstream_training}
We train Sp-HuBERT using 960 hours of LibriSpeech audio \cite{Panayotov2015_LS}, spatialised using simulated impulse responses and augmented with noise drawn from the DNS challenge dataset \cite{Reddy2021_DNS}. Unless specified otherwise, augmentation hyper-parameters are set to $p_r = 0.5$, $p_m = 0.3$, $p_n =0.5$, and the spatial loss weight $\lambda = 0.25$. We use 512 classes for the spatial loss, uniformly dividing azimuth into $m = 32$ segments, and elevation into $n = 16$ segments, resulting in an overall segmentation width of $11.25$ degrees. The Sp-HuBERT model is trained on 4 GPUs for 300k steps, with a batch size of at most 140s of audio per GPU. An Adam optimizer is used with $\beta = (0.9,0.98)$ and the learning rate ramps up linearly from zero to 3e-4 over the first 30k iterations before decaying linearly back to zero. We use the same masking configuration as HuBERT, with mask span set to 10 frames and $8\%$ of frames chosen as mask starts.

We select a value of $\lambda$ by comparing upstream validation losses. Table~\ref{tab:lambda} shows the values of the acoustic and spatial losses at 200k and 300k iterations for 3 different values of $\lambda$. It is clear from this table that increasing $\lambda$ results in a reduction in the spatial loss, with the lowest values at $\lambda=0.5$. For the acoustic loss however, we note that decreasing $\lambda$ results in diminishing returns, with only a minimal improvement from $\lambda = 0.25$ to $\lambda = 0.125$. We prioritise acoustic performance over spatial performance, as the primary purpose of the model is to achieve better performance on acoustic focused tasks in noisy environments, and therefore opt to use $\lambda=0.25$ as to minimise spatial loss without compromising on the acoustic loss.

\begin{table}[t]
    \centering
    \begin{tabular}{|c|c|c|c|c|} \hline
        $\lambda$ & \multicolumn{2}{c|}{$\mathcal{L}_\text{acoustic}$} & \multicolumn{2}{c|}{$\mathcal{L}_\text{spatial}$} \\ 
        Iters & \multicolumn{1}{c}{200k} & \multicolumn{1}{c|}{300k} & \multicolumn{1}{c}{200k} & \multicolumn{1}{c|}{200k} \\ \hline
        0.125 & 2.742 & \textbf{2.605} & 1.515 & 1.281 \\
        0.25 & \textbf{2.739} & 2.614 & 1.17 & 0.997 \\ 
        0.5 & 2.791 & 2.668 & \textbf{0.954} & \textbf{0.873} \\ \hline
    \end{tabular}
    \vspace{2mm}
    \caption{Acoustic and spatial validation losses at 200k and 300k iterations, for different values of $\lambda$}
    \label{tab:lambda}
\end{table}

\subsection{Downstream Evaluation}
We adapt a selection of tasks from various categories of the SUPERB benchmark to use both spatialisation and noise augmentation. From the speaker information category, we have chosen Speaker Identification (SID). From the content category, we have chosen Phoneme Recognition (PR) and Automatic Speech Recognition (ASR). Finally, we evaluate the Sp-HuBERT model on Emotion Recognition (ER) from the para-linguistic category. For all tasks, pre-trained upstream models are frozen, and the input to the downstream model is a trainable weighted sum of the transformer encoder layers.

For PR and ASR, we use the same task setup as the SUPERB benchmark. Both tasks are trained using a CTC loss, and performance is measured using Levenshtein distance on the phoneme sequence and word sequence respectively. The ASR task also uses the official LibriSpeech 4-gram model for language model decoding. For the SID and ER tasks, we change the downstream model from mean pooling to attentive pooling, to better accommodate the noisy setting. Both tasks are trained using a cross-entropy loss and performance is measured using classification accuracy. The four tasks are summarised in table~\ref{tab:downstream_tasks}.

For baseline comparisons, we also train downstream models for the WavLM Base and WavLM Base+ speech representations. Table \ref{tab:upstream-comparison} compares the model sizes, training times, and training set sizes of these representations to Sp-HuBERT. In terms of training time and dataset size, the closest comparison to our model is WavLM Base, while WavLM Base+ is the current state-of-the-art fully self-supervised single-channel representation model of a comparable size.

\begin{table}[t]
    \centering
    \begin{tabular}{|l|c|c|c|}
        \hline
        Model       & \#Params & Corpus & \#Iterations  \\ \hline
        WavLM Base  & 94.38M   & LS960  & 400k          \\ 
        WavLM Base+ & 94.38M   & Mix94k & 1M            \\ 
        Sp-HuBERT   & 107.39M  & LS960  & 300k          \\ \hline
        \end{tabular}
    \vspace{2mm}
    \caption{Model size, corpus, and number of training iterations for each model}
    \label{tab:upstream-comparison}
\end{table}

In addition to the acoustic tasks featured in the SUPERB Benchmark, Sp-HuBERT also learns spatial information through the spatial masked prediction loss. To evaluate the presence and accessibility of spatial information, we implement a Speech Localisation (SL) task using simulated data. The dataset is comprised of a subset of speech data taken from LibriLight \cite{Kahn2020_LL60K}, convolved with simulated FOA room impulse responses from our own dataset. Each simulated utterance contains exactly 10 seconds of audio from a stationary talker. We use a simple attention pooling downstream model for Sp-HuBERT, and train with an MSE loss on the normalised Cartesian co-ordinates of the speaker, as this was found to be the most effective method in \cite{Tang2019_FOASSL}. We measure performance using geodesic angular distance.

\begin{table}[t]
    \centering
    \begin{tabular}{|c|c|c|c|c|c|} \hline
        Task & Dataset & Model & Loss & Metric \\ \hline
        SID & Voxceleb1 & Att. Pool & CE & Acc. \\
        PR & LibriSpeech & Linear & CTC & PER \\ 
        ASR & LibriSpeech & BLSTM & CTC & WER \\
        ER & IEMOCAP & Att. Pool & CE & Acc. \\
        SL & Ours + LibriLight & Att. Pool & MSE $(x,y,z)$ & Ang. Dist \\ \hline
    \end{tabular}
    \vspace{2mm}
    \caption{Summary of the tasks for downstream evaluation}
    \label{tab:downstream_tasks}
\end{table}

Similarly to upstream training, $p_r$ controls proportion of sources that are reverberant, and $p_m$ controls the proportion of utterances that are augmented. For downstream training, we always use $p_n = 1$ so as to never augment with secondary speech.

\section{Results}\label{sec:results}
\subsection{Spatial SUPERB Benchmark Tasks}\label{sec:superb}
We train and evaluate downstream models for the SID, PR, ASR, and ER tasks in a clean setting and a noisy setting. The clean setting both trains and tests using $p_r = 0.5$, $p_m = 0$, while the noisy setting trains and tests using $p_r = 0.5$, $p_m = 1$ with mixing SNRs randomly chosen between 0 and 20dB. For all downstream tasks, we sweep over a few different learning rates and choose the model that has the best validation set performance. The learning rates used are given in Table~\ref{tab:learning_rates}.

\begin{table*}[t]
    \centering
    \begin{tabular}{|c|c|c|c|c|c|c|c|c|} \hline
        \multirow{2}{*}{Model} & \multicolumn{2}{c|}{SID} & \multicolumn{2}{c|}{PR} & \multicolumn{2}{c|}{ASR} & \multicolumn{2}{c|}{ER} \\ \cline{2-9}
                               & Clean & Noisy & Clean & Noisy & Clean & Noisy & Clean & Noisy \\ \hline
        WavLM Base & 2e-4 & 3e-4 & 2e-3 & 2e-3 & 1e-4 & 2e-4 & 1.5e-5 & 1.5e-5 \\
        WavLM Base+ & 2e-4 & 3e-4 & 2e-3 & 2e-3 & 1e-4 & 2e-4 & 1.5e-5 & 1.5e-5 \\ 
        Sp-HuBERT & 1e-4 & 2e-4 & 1e-3 & 1e-3 & 1e-4 & 1e-4 & 1.5e-5 & 1.5e-5 \\ \hline
    \end{tabular}
    \vspace{2mm}
    \caption{A summary of the learning rates used in each downstream task by each model}
    \label{tab:learning_rates}
\end{table*}

Results for Sp-HuBERT, WavLM Base, and WavLM Base+ upstream models are shown in table~\ref{tab:superb_results}. As expected, WavLM Base+ performs the best across all tasks in the clean setting due to its larger training corpus and duration, with 94000 hours of data and 1M gradient updates compared to Sp-HuBERT's 960 hours of data and 300k gradient updates. Sp-HuBERT significantly outperforms WavLM Base on Speaker ID, and shows comparable performance on ASR. In the noisy setting however, we see Sp-HuBERT offer a considerable performance improvement over both WavLM Base and Base+. With language model decoding, Sp-HuBERT achieves greater than 40\% reduction in WER when compared to WavLM Base+, along with significant improvements in SID. Across the board, the degradation in performance arising from the introduction of noise is significantly higher for WavLM Base+ when compared to Sp-HuBERT.

\begin{table*}[t]
\centering
\resizebox{\textwidth}{!}{%
\begin{tabular}{|l||c||c|cc||c||c||c|cc||c|}
\hline
\multirow{4}{*}{Model}  & \multicolumn{5}{c||}{Spatial SUPERB Clean}                                             & \multicolumn{5}{c|}{Spatial SUPERB Noisy}                                              \\ \cline{2-11} 
                        & Speaker        & \multicolumn{3}{c||}{Content}                        & ParaL          & Speaker        & \multicolumn{3}{c||}{Content}                        & ParaL          \\ \cline{2-11} 
                        & SID            & PR              & \multicolumn{2}{c||}{ASR (WER)}    & ER             & SID            & PR              & \multicolumn{2}{c||}{ASR (WER)}    & ER             \\ \cline{2-11} 
                        & Acc.$\uparrow$ & PER$\downarrow$ & LM$\downarrow$ & No LM$\downarrow$ & Acc.$\uparrow$ & Acc.$\uparrow$ & PER$\downarrow$ & LM$\downarrow$ & No LM$\downarrow$ & Acc.$\uparrow$ \\ \hline
WavLM Base              & 62.51          & 6.43            & 5.82           & 7.83              & 59.00          & 54.08          & 17.85           & 18.04          & 20.43             & 55.05          \\ \hline
WavLM Base+             & \textbf{77.03} & \textbf{5.06}   & \textbf{4.78}  & \textbf{6.56}     & \textbf{61.74} & 65.48          & 13.62           & 13.26          & 15.26             & 58.79          \\ \hline
Sp-HuBERT               & 73.10          & 7.25            & 5.70           & 7.87              & 60.86          & \textbf{69.43} & \textbf{9.58}   & \textbf{7.84}  & \textbf{10.46}    & \textbf{59.77} \\ \hline
\end{tabular}}
\vspace{2mm}
\caption{Results of WavLM Base, WavLM Base+ and SpHuBERT on a spatial version of 4 tasks from the SUPERB benchmark}
\label{tab:superb_results}
\end{table*}

\subsection{Sensitivity to Noise}\label{sec:noise_sensitivity}
\begin{figure*}[t]
\centering
\subfloat[Phoneme Error Rate vs SNR (lower is better)]{\includegraphics[width=3.2in]{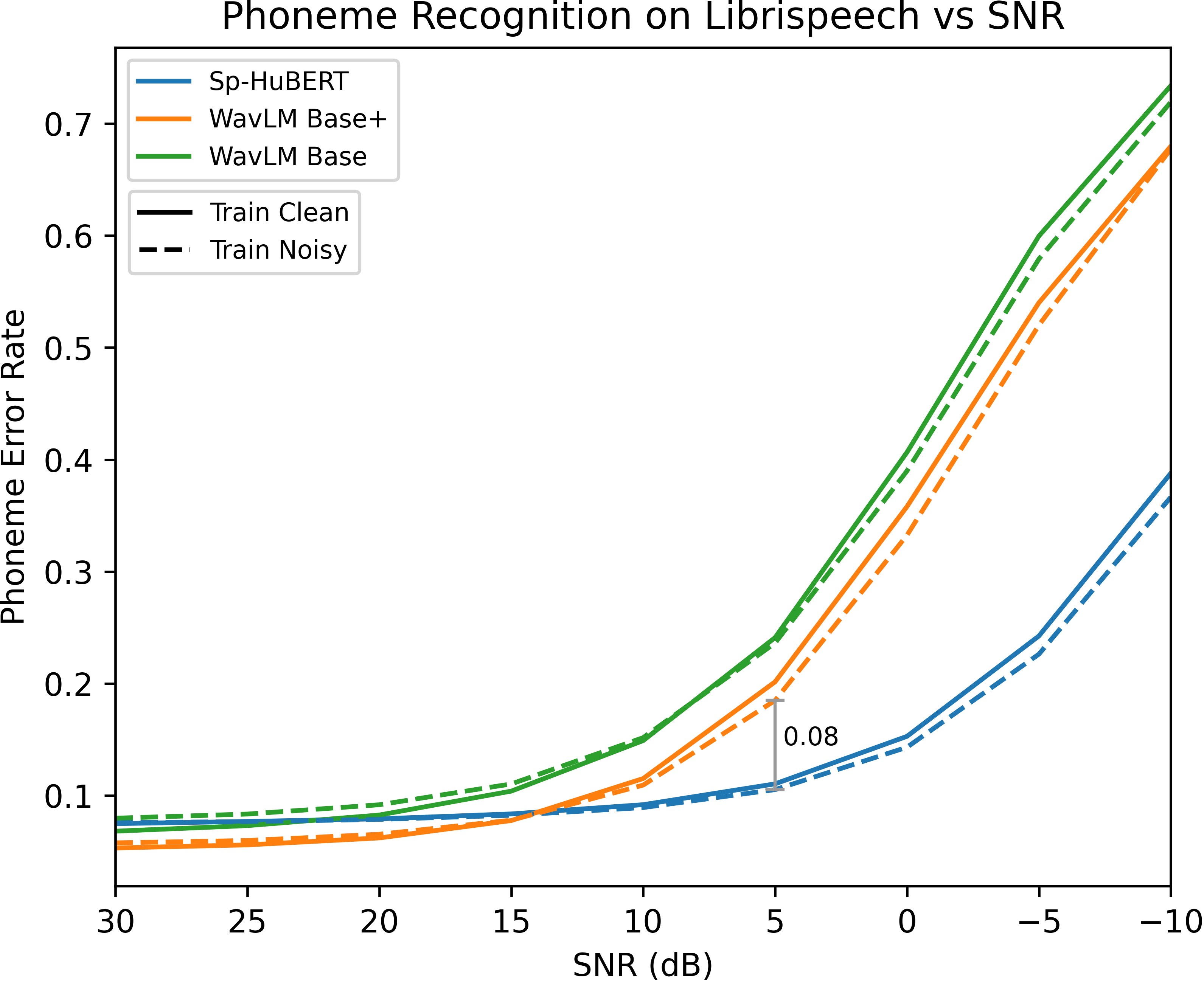}%
\label{fig:libriphone_per_vs_snr_train}}
\hfil
\subfloat[Classification Accuracy vs SNR (higher is better)]{\includegraphics[width=3.2in]{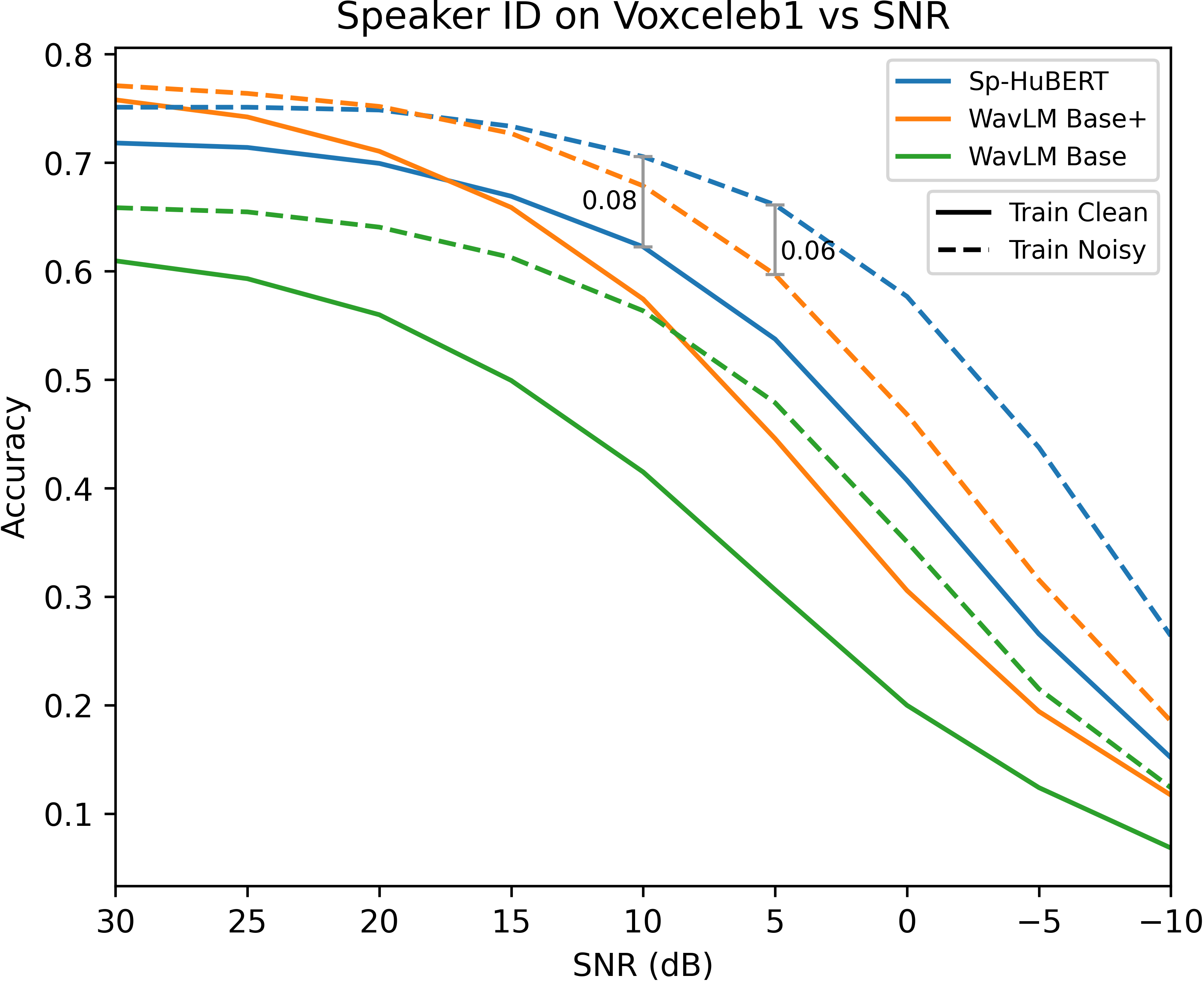}%
\label{fig:voxceleb_acc_vs_snr_train}}
\caption{A performance comparison between Sphubert, WavLM Base+ and WavLM Base at various SNRs for two tasks. Solid lines show performance when the downstream model is trained only on clean speech, and dashed lines show performance when the downstream model is trained on noisy speech of SNRs varying from 0dB to 20dB.}
\label{fig:err_vs_snr_train}
\end{figure*}

Figure~\ref{fig:err_vs_snr_train} shows the performance of each upstream model vs SNR on the PR and SID tasks. The solid lines show performance using the downstream model trained only on clean speech, while the dashed lines show performance using the downstream model trained with noise at 0-20dB SNR. On both tasks, Sp-HuBERT begins to outperform WavLM Base+ when the SNR drops below 15dB. At 5dB, Sp-HuBERT achieves an 8\% reduction in phoneme error rate on Librispeech, and a 6\% improvement in classification accuracy on Voxceleb1.

The difference between performance when training the downstream model on noisy data is another key point of interest here. We observe that on the phoneme recognition task, exposing the downstream model to noise during training has a minimal impact on performance, but on the speaker identification task, there is a significant improvement gained by training on noisy data, with an 8\% increase in absolute accuracy at 10dB SNR when using Sp-HuBERT.

This difference in performance indicates that when exposed to noise during training, the downstream model is able to learn a more effective way to extract speaker information from the representation model. The mechanism behind this effect will be discussed further in section~\ref{sec:layer_weight_analysis}.

\subsection{Sensitivity to Reverb}
\begin{figure*}[!t]
\centering
\subfloat[Word Error Rate vs SNR (lower is better)]{\includegraphics[width=3.2in]{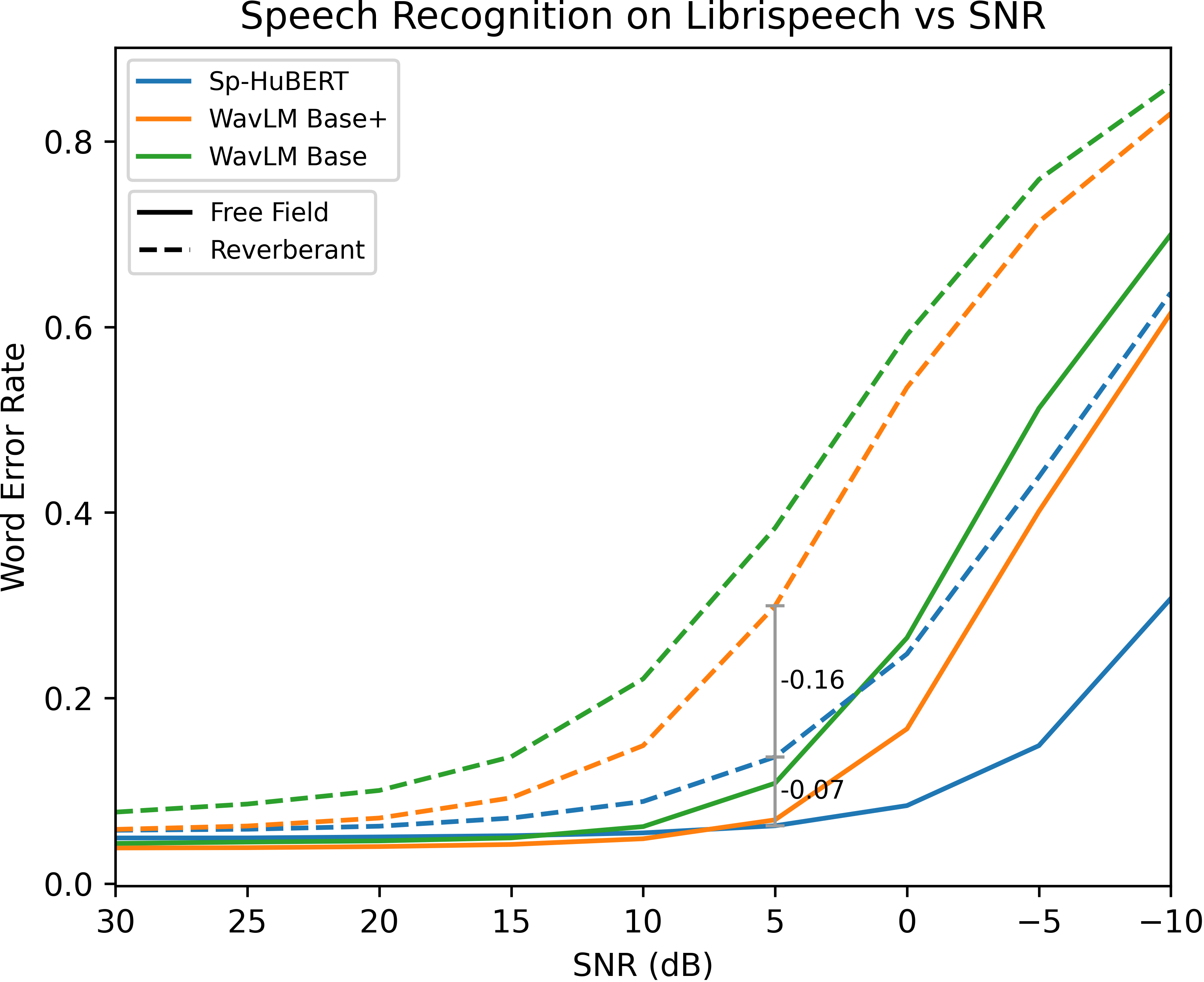}%
\label{fig:librispeech_wer_vs_snr_reverb}}
\hfil
\subfloat[Classification Accuracy vs SNR (higher is better)]{\includegraphics[width=3.2in]{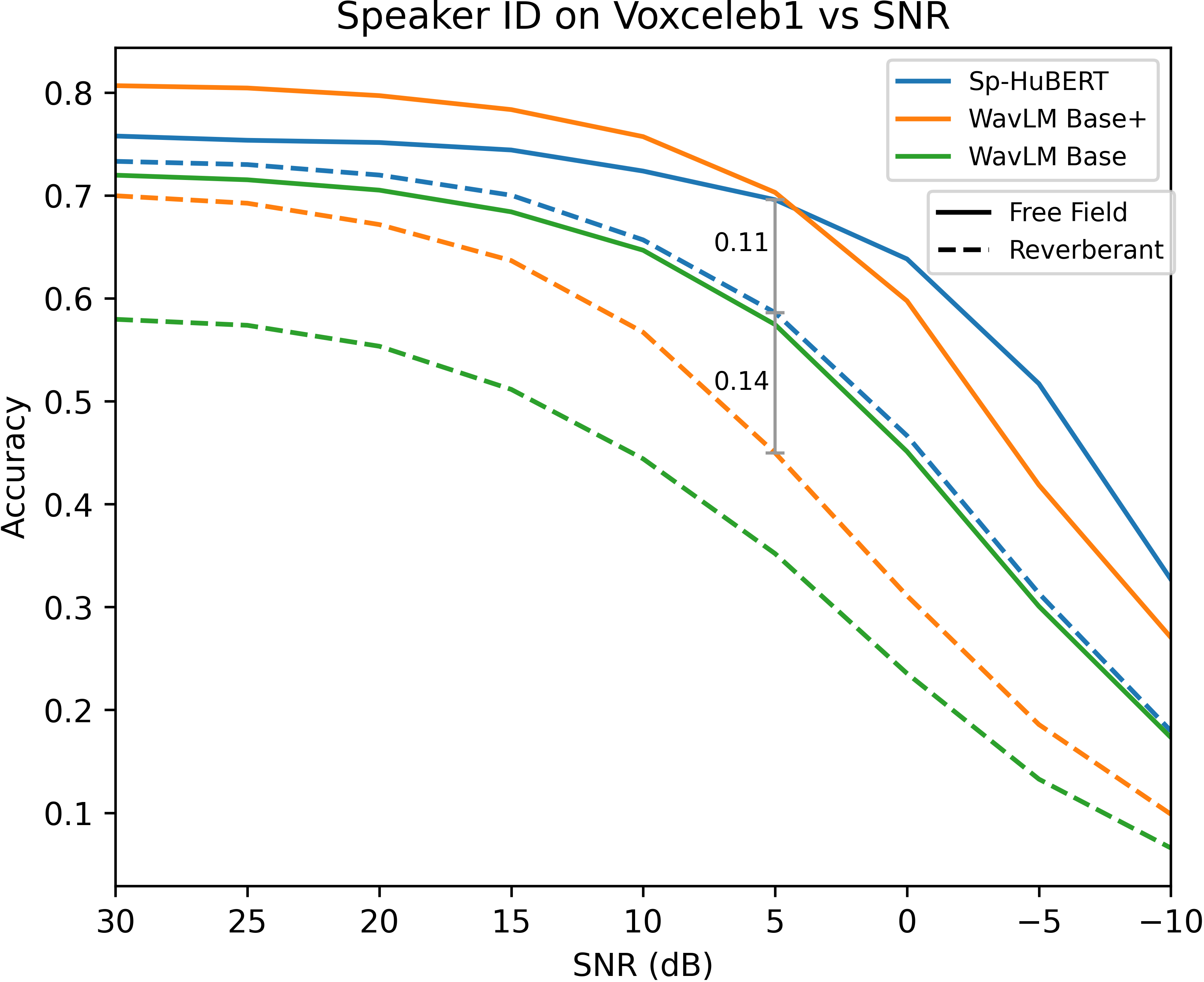}%
\label{fig:voxceleb_acc_vs_snr_reverb}}
\caption{A performance comparison between Sphubert, WavLM Base+ and WavLM Base at various SNRs for ASR on Librispeech and Speaker Identification on Voxceleb1. Solid lines show performance when on free field signals, and dashed lines show performance on reverberant signals.}
\label{fig:err_vs_snr_reverb}
\end{figure*}

Figure~\ref{fig:err_vs_snr_reverb} shows the performance of each upstream model vs SNR on the ASR and SID tasks with different reverberation conditions, using the downstream model trained on noisy data. Dashed lines show the performance on test data with both reverberant speech and noise, while solid lines show the performance on free field speech and noise mixtures.

On both tasks, reverberation has a significant impact on the performance of the representations. For Sp-HuBERT, WER increases by 7\% and SID accuracy decreases by 11\% at 5dB SNR when introducing reverberation. However, the performance degradation is more severe for WavLM. On both tasks, even at high SNR Sp-HuBERT outperforms WavLM Base+ in reverberant conditions. Particularly on the ASR task, the performance of Sp-HuBERT degrades significantly slower than that of WavLM as the SNR decreases, with Sp-HuBERT offering a 16\% WER improvement at 5dB in reverberant conditions.

\subsection{Speech Localisation}
Similarly to the speech tasks in section~\ref{sec:superb}, we train two downstream models to solve the Speech Localisation task. The clean trained model uses $p_r = 1, p_m = 0$, and the noisy trained model uses $p_r = 1, p_m = 1$ with random SNRs randomly sampled between 0 and 20dB. We evaluate the performance of both models in free-field and reverberant settings. We do not compare to baseline representations for this task, as this is the first work to produce a spatial representation.

\begin{figure}[t]
    \centering
    \includegraphics[width=3.2in]{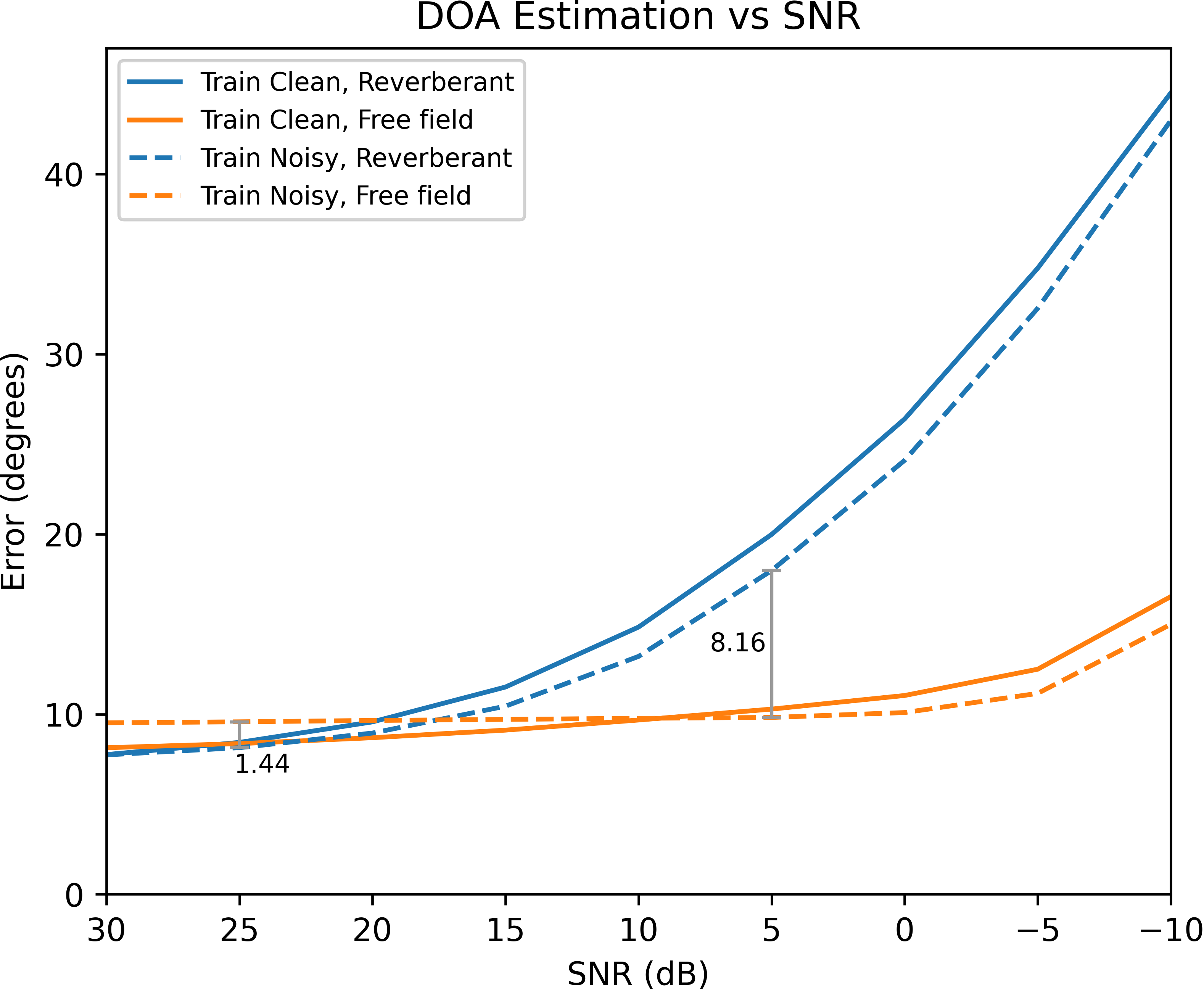}
    \caption{Speech Localisation performance vs SNR for Sp-HuBERT with downstream models trained on clean and noisy data, in both free field and reverberant conditions.}
    \label{fig:doa_estimation}
\end{figure}

Figure~\ref{fig:doa_estimation} shows angular error vs SNR of both models in reverberant and free field testing scenarios. Firstly, we see that as SNR decreases, the presence of reverb significantly increases the difficulty of the task. On free field recordings, the performance at 5dB SNR is nearly the same as the performance at 30dB SNR, while in the reverberant recordings the average angular error increases by over 8 degrees. At high SNRs however, the model appears to perform better under reverberant conditions. This is partially due to the fact that the downstream models were both trained with $p_r = 1$.

Next we compare training on clean data to training on noisy data. Firstly, we see that in reverberant environments, the noisy trained model consistently performs better than the clean trained version. In the free field test case however, we find that the model trained on clean data performs better at high SNRs, most likely due to these conditions more closely matching their training data. 

We note that at high SNRs, localisation in free-field conditions on clean speech is a simple task in which traditional methods can easily obtain very high accuracy, but Sp-HuBERT averages around 8 degrees error at 30dB SNR. This is a significant limitation of the upstream model caused by the quantisation used during training, which separates both azimuth and elevation into segments with a width of 11.25 degrees. We hypothesise that using discrete DOA labels for upstream training restricts the resolution of the spatial information in the representation.

\subsection{Layer Weight Analysis}\label{sec:layer_weight_analysis}
\begin{figure*}[!t]
\centering
\subfloat[Sp-HuBERT, Clean Featuriser]{\includegraphics[width=3.0in]{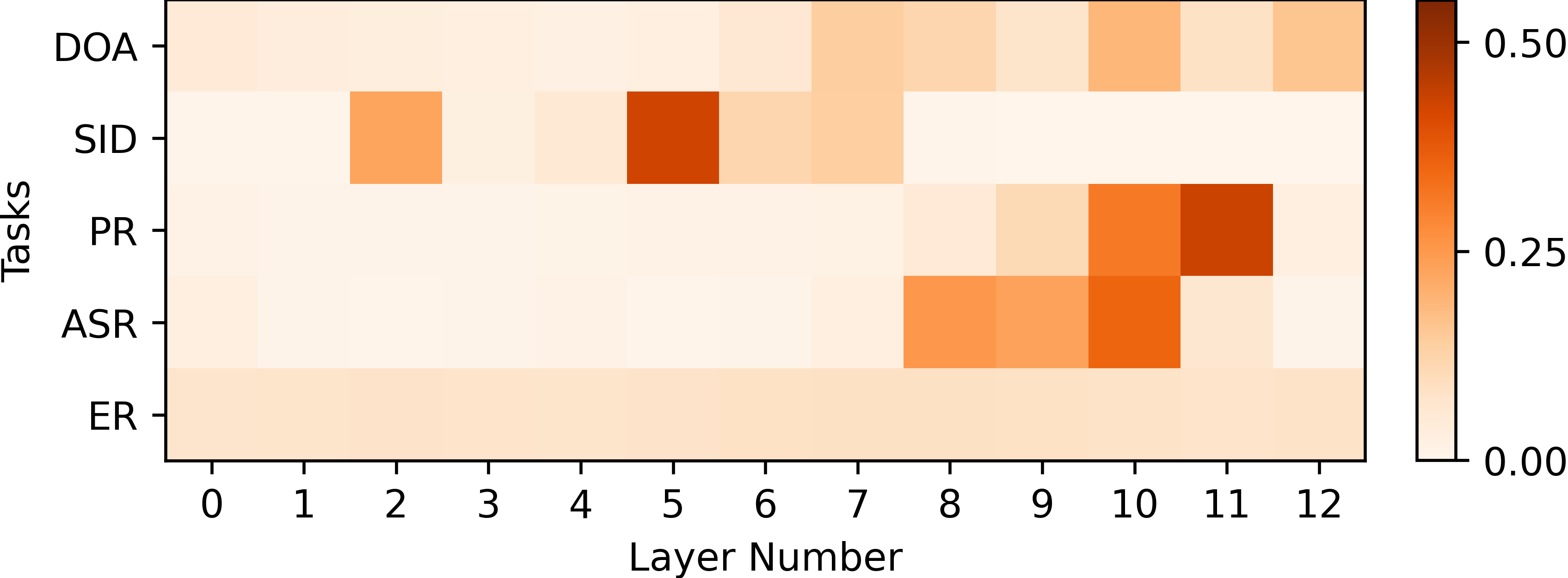}%
\label{fig:sp-hubert_clean_weights}}
\hfil
\subfloat[WavLM Base+, Clean Featuriser]{\includegraphics[width=3.0in]{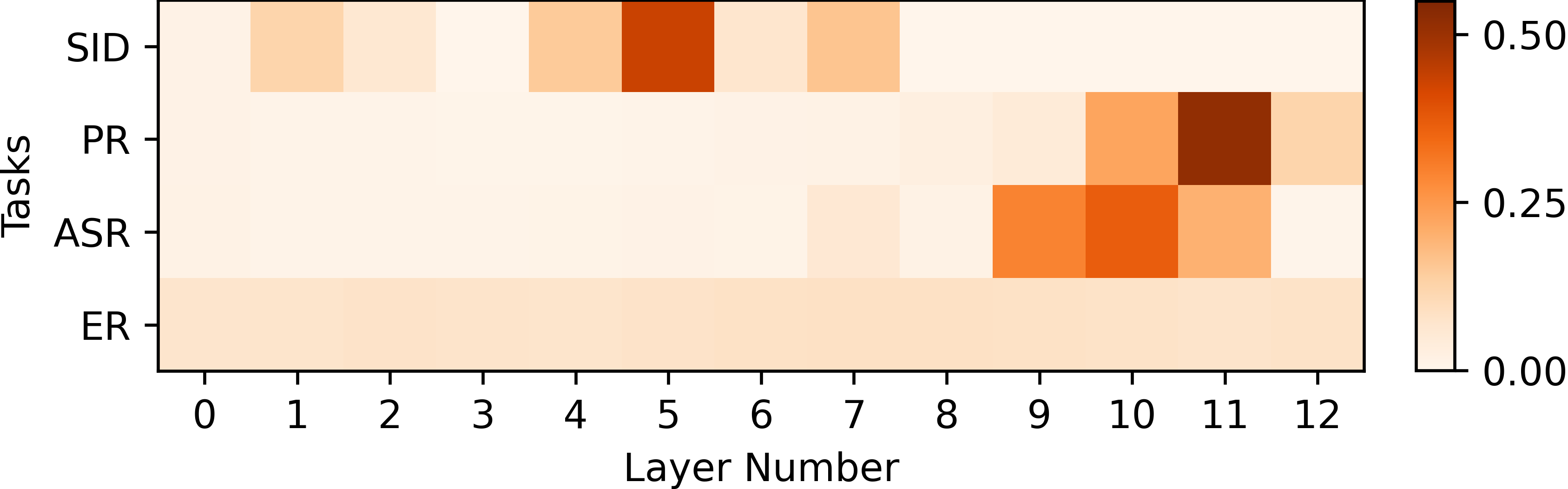}%
\label{fig:wavlm_clean_weights}} 
\\
\subfloat[Sp-HuBERT, Noisy Featuriser]{\includegraphics[width=3.0in]{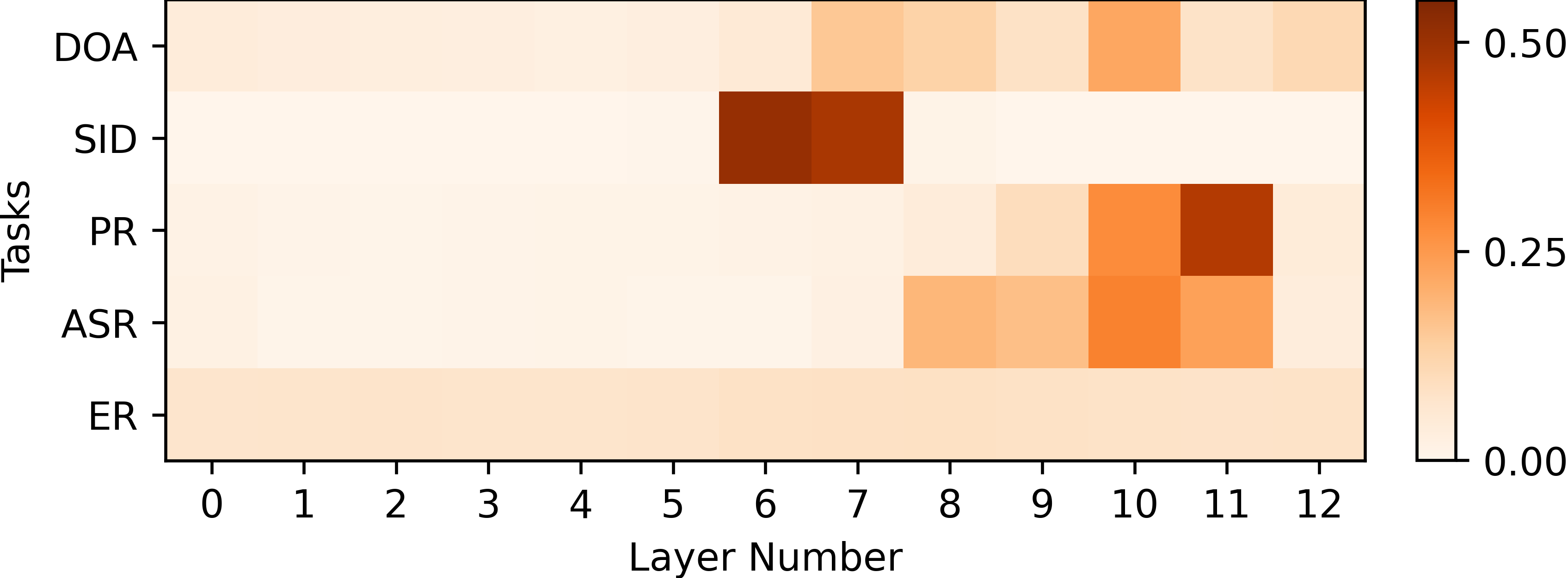}%
\label{fig:sp-hubert_noisy_weights}}
\hfil
\subfloat[WavLM Base+, Noisy Featuriser]{\includegraphics[width=3.0in]{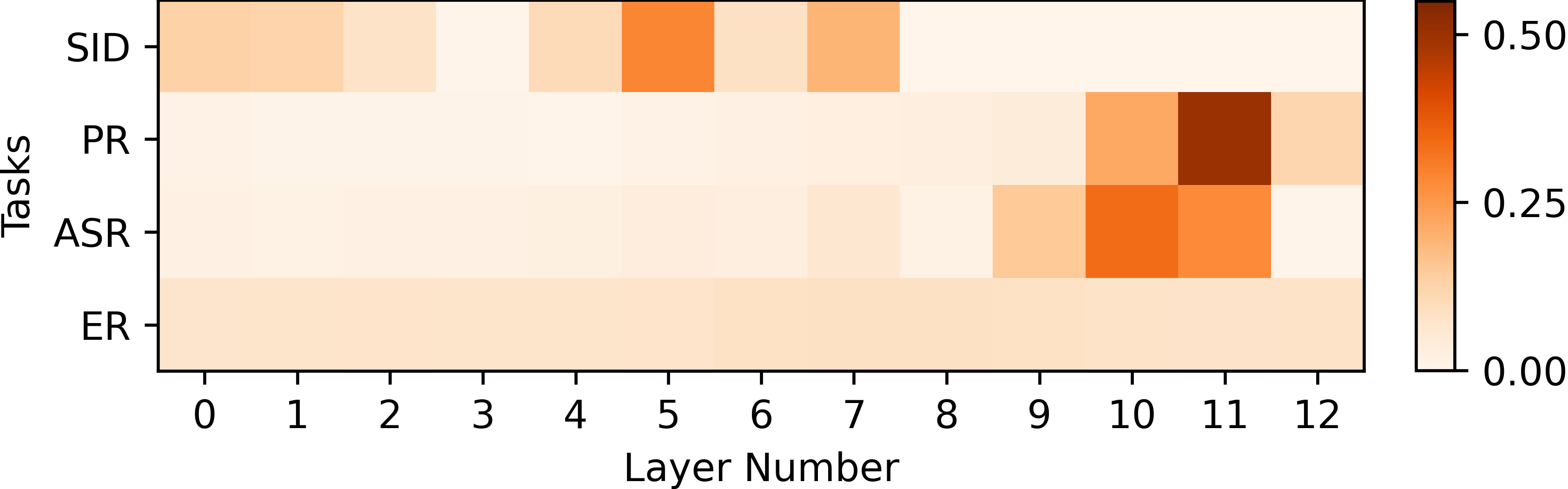}%
\label{fig:wavlm_noisy_weights}}
\\
\subfloat[Sp-HuBERT Weight Difference]{\includegraphics[width=3.0in]{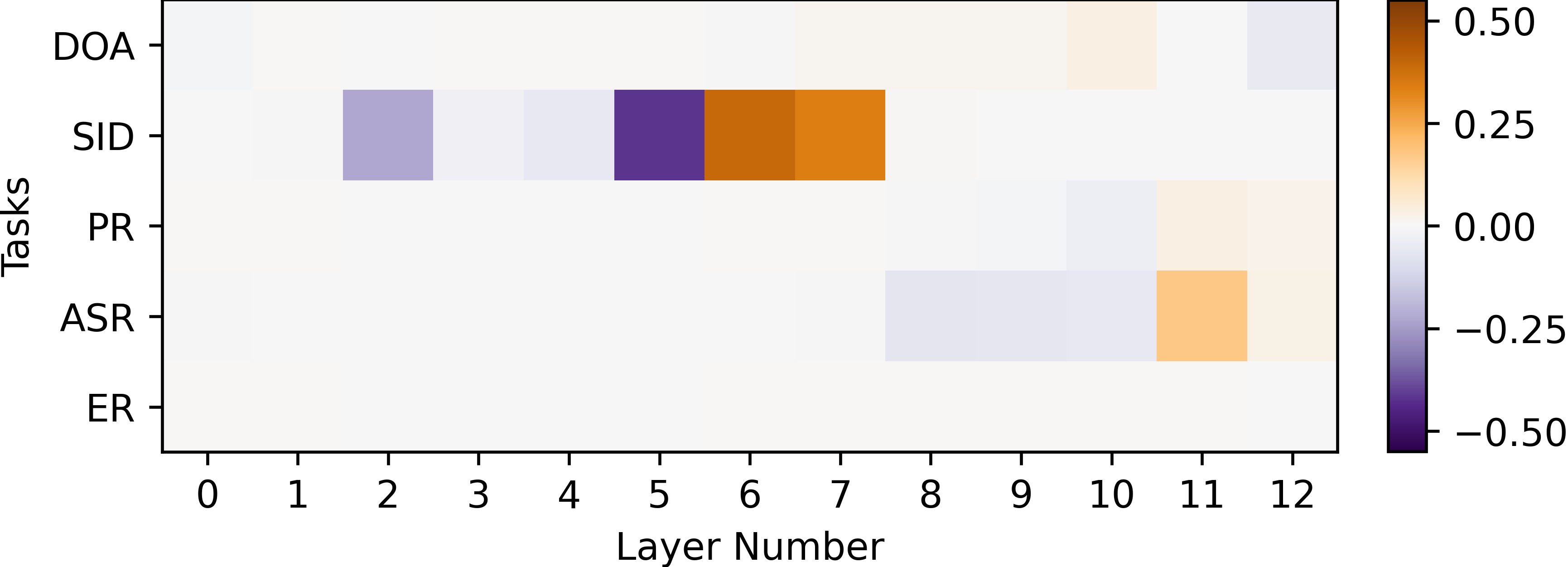}%
\label{fig:sp-hubert_weight_difference}}
\hfil
\subfloat[WavLM Base+ Weight Difference]{\includegraphics[width=3.0in]{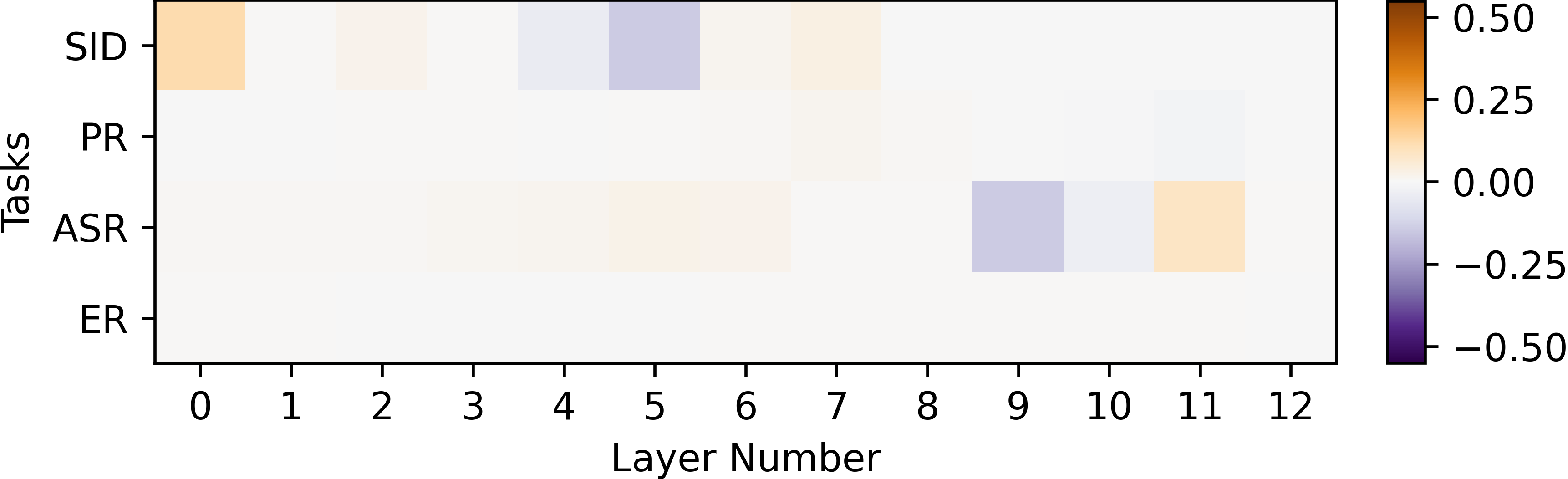}%
\label{fig:wavlm_weight_difference}}
\caption{Featuriser weight breakdown for Sp-HuBERT and WavLM Base+ for each of the tasks tested in the benchmark along with Speech Localisation for Sp-HuBERT. Layer 0 corresponds to the input to the transformer encoder. The y-axis represents different tasks, and the x-axis represents the weight given to each layer.}
\label{fig:weight_analysis}
\end{figure*}

Following the approach of \cite{Chen2021_WavLM}, we investigate the contribution of each layer of the transformer encoder to each of the 4 downstream tasks along with Speech Localisation for Sp-HuBERT. The input to each downstream model that we trained in our earlier experiments is a weighted-sum of the 13 layers of the transformer encoder, including the input layer. These weights indicate which layers provide the most information for the downstream models in each task. Figure~\ref{fig:weight_analysis} shows the weights learned for each task, both when trained on clean spatial speech and when trained on noisy data. Larger layer weights indicate greater contribution of the corresponding layer.

Figures~\ref{fig:sp-hubert_clean_weights} and \ref{fig:wavlm_clean_weights} show that the weights learned for each of the 4 tasks are similar in both Sp-HuBERT and WavLM when trained on clean data. Consistent with the findings of \cite{Chen2021_WavLM, Chen2022_SSRASV}, we see that speaker information is most easily accessible from the earlier layers of the model, with the dominant weight at layer 5, while phoneme recognition and automatic speech recognition utilise layers closer to the end of the model. We also note that the layer weights for emotion recognition are near uniform, with all layers contributing very similar amounts. For the SL task, once again there is an increased contribution in the later layers, particularly layers 10 and 12.

Figures~\ref{fig:sp-hubert_noisy_weights} and \ref{fig:wavlm_noisy_weights} show the weights learned when trained on noisy data for both Sp-HuBERT and WavLM, while figures~\ref{fig:sp-hubert_weight_difference} and \ref{fig:wavlm_weight_difference} show the difference between the weights trained on noisy and clean data. For WavLM there are some subtle changes between the clean and noisy case, with an increase in the use of layer 0 for SID and an increase in the use of layer 11 for ASR. In contrast, there is a significant change in weights for Sp-HuBERT. For the SID task, the downstream model trained on noisy data is using layers 6 and 7 almost exclusively, indicating that the speaker information in these layers is far more robust to noise than that in layer 5. We also see a slight preference towards deeper layers in the ASR task, with a notable increase in the weight of layer 11 and a decrease in the weights of layers 8-10. This suggests that particularly in the case of Sp-HuBERT, later layers of the representation tend to be more robust to spatial noise than earlier layers.

This analysis also provides some insight on the performance improvements when training on noise that were previously observed in section~\ref{sec:noise_sensitivity}. Through the layer weights, we see a clear difference in how the two downstream models extract information from the representations in each of the tasks. For both Sp-HuBERT and WavLM Base+, we see the most significant changes in layer weights between clean and noisy on the SID task, on which a substantial performance improvement was observed. In contrast, we see minimal change in layer weights on the PR task, on which only minimal performance improvements were observed. It appears that exposing the downstream model to noise during training allows it to select layers of the representation that contain the required speaker information, and are more robust to the noise sources. In the case of phonetic information however, it appears that no significant advantages can be found in other layers.

\section{Conclusion}
This paper presents Spatial HuBERT, a self-supervised spatial speech representation model trained on a spatial speech dataset generated using simulated first order ambisonics impulse responses, which we release to the public for future development. Spatial HuBERT extends the masked prediction and denoising losses of HuBERT and WavLM with a spatial loss term and produces representations that are more robust to both noise and reverberation than state-of-the-art single channel models. Despite training on only 960 hours of data from LibriSpeech, Spatial HuBERT outperforms even WavLM Base+ on a variety of downstream tasks in noisy testing conditions. Additionally, the representations learned by Spatial HuBERT contain spatial information, enabling its use for speech localisation tasks.

For future work, we aim to increase the size of the training corpus and scale up the size of the model to enable comparisons with WavLM Large. Another potential avenue for improvement involves incorporating the loss terms from Cocktail HuBERT \cite{Fazel-Zarandi2023_cocktailhubert}, to train the model to disentangle multiple simultaneous talkers in noisy spatial environments.

\section*{Acknowledgments}
The work for this paper was conducted as a Research Internship at Dolby Australia. We thank them for providing the compute resources required to conduct our experiments. We also thank Henry Chen and David McGrath for useful discussions, and Dylan Harper-Harris for his assistance in debugging our code.

\bibliographystyle{IEEEtran}
\bibliography{refs}

\newpage

\vfill

\end{document}